\documentclass[11pt]{article}

\usepackage{amsmath}
\usepackage{amssymb}
\usepackage{booktabs}
\usepackage{graphicx}
\usepackage[colorlinks=true,linkcolor=blue,citecolor=blue,urlcolor=blue]{hyperref}
\usepackage[round]{natbib}
\usepackage[margin=1in]{geometry}
\usepackage{xcolor}
\usepackage[section]{placeins}
\usepackage[font=small,labelfont=bf]{caption}

\title{From Topic to Transition Structure: Unsupervised Concept Discovery\\at Corpus Scale via Predictive Associative Memory}
\author{Jason Dury\\Independent Researcher\\jason@eridos.ai}
\date{}

\begin{document}

\maketitle

\begin{abstract}
\noindent Embedding models group text passages by semantic content or what text is \emph{about}. We show that a different training signal, temporal co-occurrence within texts, discovers a different kind of structure: recurrent transition-structure concepts---what text \emph{does}. We train a 29.4M-parameter contrastive model on 373 million co-occurrence pairs extracted from 9,766 Project Gutenberg texts (24.96 million passages), mapping pre-trained embeddings into an association space where passages with similar local transition structure cluster together. Under capacity constraint (42.75\% training accuracy), the model cannot simply memorise the full co-occurrence set and instead must compress across recurring patterns. Clustering at six granularities ($k{=}50$ to $k{=}2{,}000$) produces a multi-resolution map of textual concepts---from broad narrative modes like ``direct confrontation'' and ``lyrical meditation'' to precise registers, literary traditions, and scene templates like ``Franco-Prussian War military dispatches,'' ``sailor dialect,'' and ``courtroom cross-examination.'' At $k{=}100$, clusters average 4,508 books each (of 9,766), confirming these are corpus-wide patterns, not author-specific artefacts. Direct comparison with embedding-similarity clustering (BGE-large-en-v1.5) shows that raw embeddings group by topic while association-space clusters group by function, register, and literary tradition. Novels not seen during training are assigned to existing clusters without retraining; the association model concentrates each novel into a selective subset of structurally coherent clusters, while raw embedding assignment saturates nearly all clusters. Validation controls reduce the likelihood of positional, length, and book-concentration confounds. The method extends Predictive Associative Memory (PAM) from episodic recall to concept formation: where PAM recalls specific associations and inductive transfer fails, multi-epoch contrastive training under compression extracts structural patterns that transfer to unseen texts---the same framework producing qualitatively different behaviour in a different regime.
\end{abstract}

\section{Introduction}
\label{sec:introduction}

Consider two passages. In the first, a narrator in a South American rainforest is seized by ``unspeakable horror'' and ``remorse,'' starting up ``with a cry of anguish.'' In the second, a guest in an English country house lies awake, unable to shake ``that odious detail'' of a still, dark figure waiting on the landing. The settings share nothing: tropical forest versus Victorian manor, physical violence versus quiet dread, first-person confession versus third-person observation. No topic model would group them. No keyword search would connect them. Yet both passages perform the same narrative work---they are the moment a character confronts something that should not be there, the gear-shift from ordinary experience into horror.

This structural kinship is invisible to similarity-based text analysis. Embedding models trained on semantic content will group all horror passages together, all rainforest passages together, all Victorian domestic passages together---sorting by \emph{what text is about}. What they miss is \emph{what text does}: its role in the local structure of a narrative. A passage describing a moonlit face through a window may function as romantic anticipation in one novel and supernatural dread in another. The semantic content is similar; the structural role is opposite.

We show that temporal co-occurrence within texts---which passages tend to appear near which other passages---provides a training signal that discovers this kind of structure. The principle is drawn from Predictive Associative Memory (PAM) \citep{dury2026pam}: useful relationships often connect items that were \emph{experienced together} rather than items that \emph{appear similar}. In sequential text, passages that reliably co-occur within a local window share transition structure---similar predecessors, similar successors, similar narrative neighbourhoods---even when their vocabulary, genre, and century of composition differ entirely.

We train a small contrastive model (29.4M parameters) on temporal co-occurrence pairs extracted from 9,766 Project Gutenberg texts (24.96 million passages). The model maps pre-trained text embeddings into an association space where passages with similar transition structure cluster together. Under a capacity bottleneck---the model reaches only 42.75\% training accuracy across 373 million pairs---it cannot simply memorise the full co-occurrence set and is instead forced to compress across the shared patterns that recur across thousands of books. The resulting clusters look nothing like topics. They are \emph{concepts}: cross-author, cross-genre patterns of transition structure that emerge from the statistics of what-follows-what. Some correspond to narrative functions---confrontation, investigation, departure. Others capture discourse registers, literary traditions, scene templates, or recurring subject-matter conventions. The common thread is that they are organised by \emph{what text does in context} rather than by what it says.

At coarse resolution ($k{=}50$ clusters), these concepts capture broad modes: direct confrontation, lyrical meditation, domestic routine, historical combat. At fine resolution ($k{=}2{,}000$), they decompose into precise variants: Franco-Prussian War military dispatches, Galsworthy-era social realism, Victorian book advertisements, sailor dialect, courtroom cross-examination scenes. Multi-resolution clustering from $k{=}50$ to $k{=}2{,}000$ reveals hierarchical structure on inspection---zooming in is analogous to zooming from ``action sequence'' to ``chase scene on horseback.'' At every level, clusters draw from hundreds to thousands of different books across genres and centuries, confirming that the discovered structure is not an artefact of any single author or tradition.

Three lines of evidence distinguish these concepts from what standard text embeddings produce. First, direct comparison: clustering the same corpus by raw embedding similarity (BGE-large-en-v1.5) produces topical groupings---all fear passages together, all money passages together---while the association-space clusters group by transition structure: narrative function, discourse register, and literary tradition (stage plays, epistolary prose, cynical worldly wisdom, sailor dialect). Second, unseen-novel evaluation: five canonical novels absent from the training corpus are processed through the trained model and assigned to existing clusters without retraining. The association model concentrates each novel into a selective subset of clusters that track structural role and shift at mode boundaries; raw embedding assignment saturates nearly all clusters, tracking topic rather than function. Third, validation controls reduce the likelihood of position-in-book, token count, and book concentration confounds, while a temporal shuffle control on the pilot corpus collapses cross-boundary recall by 95.2\%, confirming the signal is genuine co-occurrence structure.

This work sits between two prior results. The PAM framework \citep{dury2026pam} established that temporal co-occurrence trains predictors capable of faithful associative recall across representational boundaries, validated on a synthetic benchmark. Association-Augmented Retrieval (AAR) \citep{dury2026aar} applied the same principle to multi-hop passage retrieval, demonstrating +8.6 Recall@5 on HotpotQA---but showed that inductive transfer fails, because passage-to-passage associations are contingent. The present work occupies a different regime: instead of recalling specific associations, the model extracts \emph{structural patterns} that compress across thousands of texts. Multi-epoch contrastive training under capacity constraint acts as a consolidation mechanism---analogous to hippocampal replay compressing episodes into stable representations---and the resulting concepts transfer inductively to unseen texts precisely because they capture recurrent structure rather than contingent co-occurrences.

\textbf{Contributions:}

\begin{enumerate}
    \item A method for unsupervised concept discovery from temporal co-occurrence in text, requiring no labels, no topic annotations, and no genre metadata. The training signal is which passages appear near which other passages; the concepts emerge from compression.
    \item Empirical demonstration that the discovered concepts capture transition structure---narrative function, discourse register, literary tradition---rather than topic, validated by direct comparison with embedding-similarity clustering, unseen-novel evaluation, and confound controls.
    \item A multi-resolution concept map at six granularities ($k{=}50$ to $k{=}2{,}000$) over 9,766 texts, with an interactive demonstration tool for exploration.
    \item Evidence that the PAM framework's association $\neq$ similarity principle, previously validated for episodic recall and multi-hop retrieval, extends to concept discovery at corpus scale---with qualitatively different transfer properties (inductive transfer succeeds here, unlike in retrieval).
\end{enumerate}

\section{Related Work}
\label{sec:related}

\subsection{Topic Modelling}
\label{sec:topic}

Latent Dirichlet Allocation (LDA) \citep{blei2003lda} discovers topics as distributions over words, assigning documents mixtures of topics. Neural topic models and BERTopic \citep{grootendorst2022bertopic} extend this with pre-trained embeddings, producing more coherent topics at the cost of increased complexity. All topic models share a fundamental property: they group text by \emph{content}---which words or embedding regions co-occur within the same document or passage. A topic model applied to our corpus would discover topics like ``military vocabulary,'' ``romantic language,'' or ``nautical terminology.'' What it cannot discover is that a passage of military vocabulary and a passage of romantic language may serve the same narrative function---the moment of departure---if they reliably occupy the same structural position across different novels.

\subsection{Computational Narrative Analysis}
\label{sec:narrative}

Computational approaches to narrative structure have largely focused on labelled frameworks. Story grammars and plot unit extraction \citep{chambers2008narrative} require annotated schemas. Sentiment arc analysis \citep{reagan2016arcs, jockers2015syuzhet} captures emotional trajectory but reduces narrative to a single valence dimension. Character network analysis \citep{elson2010social} captures relational structure but not passage-level narrative function. Recent work on narrative event chains \citep{chambers2008narrative, chambers2009schemas} learns typical event sequences but operates at the event level, not the passage level, and requires event extraction pipelines.

The closest precedent is work on narrative functions in folklore and mythology. Propp's morphology of the folktale \citep{propp1968morphology} identified 31 narrative functions (departure, interdiction, violation, etc.) that recur across Russian fairy tales regardless of surface content---precisely the kind of structure our method discovers, but from temporal statistics rather than manual annotation. Our contribution is a mechanism for learning such functions from data at scale.

\subsection{Contrastive Representation Learning}
\label{sec:contrastive}

Contrastive learning trains representations by pulling associated items together and pushing non-associated items apart \citep{oord2018cpc, radford2021clip}. In NLP, contrastive methods have been applied to sentence embeddings \citep{gao2021simcse}, document representations \citep{izacard2022contriever}, and passage retrieval \citep{karpukhin2020dpr}. These methods typically define positives through semantic equivalence (paraphrase, entailment) or task relevance (query-document pairs). Our training signal is different: positives are passages that co-occur within a temporal window in the same text, regardless of semantic similarity. This produces a fundamentally different embedding geometry---one organised by transition structure rather than content.

\subsection{Predictive Associative Memory and Association-Augmented Retrieval}
\label{sec:pam-aar}

The Predictive Associative Memory framework \citep{dury2026pam} formalises the distinction between similarity-based and association-based retrieval. A JEPA-style predictor trained on temporal co-occurrence learns to navigate associative structure in embedding space, retrieving items that co-occurred with a query regardless of their representational similarity. On a synthetic benchmark, this predictor achieves 97\% precision at rank 1 for temporal associates, while cosine similarity scores zero on cross-boundary pairs. A temporal shuffle control confirms the signal is genuine co-occurrence structure.

AAR \citep{dury2026aar} operationalised this principle for multi-hop passage retrieval, training a contrastive MLP on passage co-occurrence annotations and demonstrating +8.6 Recall@5 on HotpotQA. Critically, AAR's associations are corpus-specific by design: a model trained on one set of passage co-occurrences does not generalise to unseen pairs, confirming that the learned associations are faithful to experienced co-occurrences rather than abstract patterns.

The present work shares architecture with both: a contrastive MLP trained on temporal co-occurrence, operating on pre-trained embeddings. The main difference lies in the evaluation target. PAM tests faithful recall of specific associations. AAR tests retrieval improvement on specific corpora. Here, we test \emph{concept discovery}---whether cross-author structural patterns emerge under compression. AAR's inductive transfer fails because its associations are contingent; ours succeeds because the patterns are recurrent. Same framework, different compression regime, different behaviour.

\section{Method}
\label{sec:method}

\begin{figure}[!htbp]
\centering
\includegraphics[width=\textwidth]{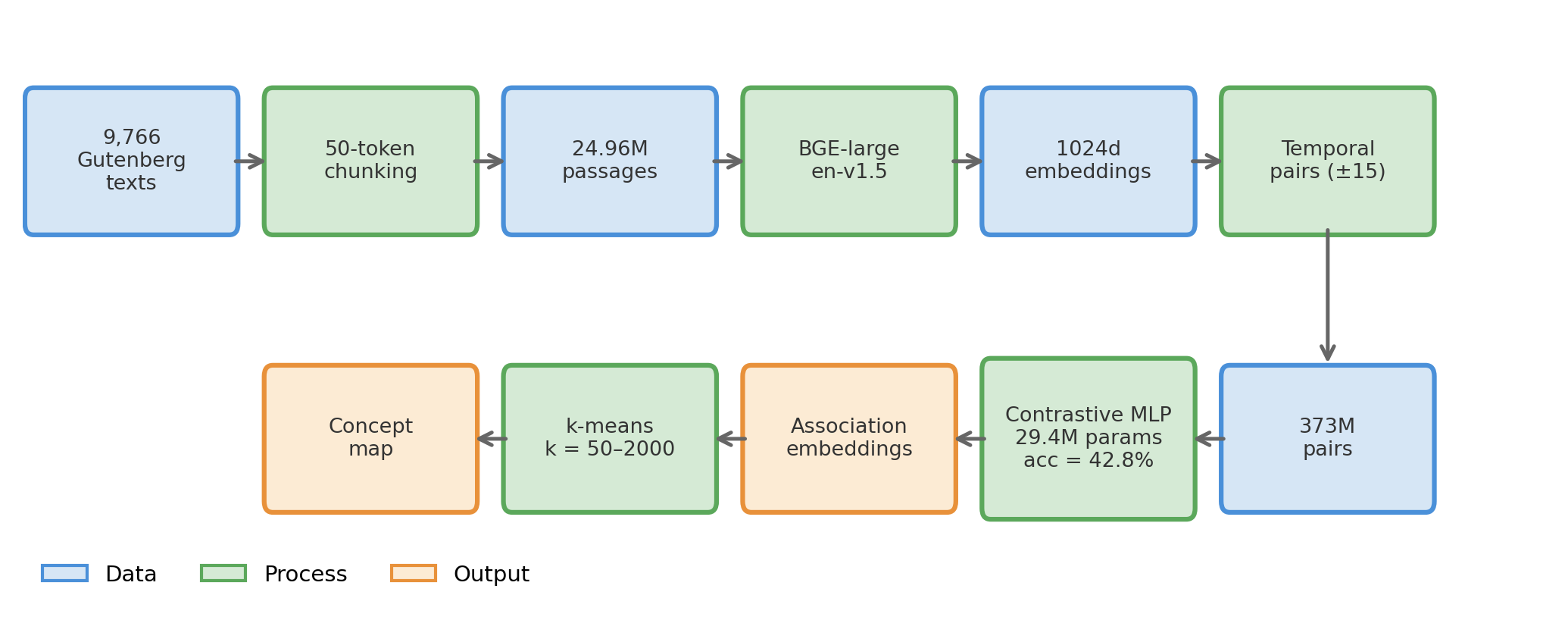}
\caption{Method overview. Texts are chunked into short passages, embedded with BGE-large-en-v1.5, and temporal co-occurrence pairs are extracted within a 15-chunk window. A contrastive MLP maps embeddings into association space; $k$-means clustering at six granularities ($k{=}50$ to $k{=}2{,}000$) produces a multi-resolution concept map.}
\label{fig:pipeline}
\end{figure}

The method (Figure~\ref{fig:pipeline}) proceeds in four stages: chunking texts into short passages, embedding them, training a contrastive model on temporal co-occurrence pairs, and clustering the transformed embeddings at multiple resolutions.

\subsection{Corpus and Chunking}
\label{sec:chunking}

We use 10,000 English-language texts from Project Gutenberg \citep{gutenberg}, retrieved via the Gutendex API \citep{gutendex} in ascending Gutenberg ID order (IDs up to approximately 13,700). After excluding 234 texts that fail chunking (malformed encoding, insufficient length, or non-prose content), the final corpus comprises 9,766 texts spanning fiction, non-fiction, essays, drama, poetry collections, and religious texts from the 16th to early 20th centuries. The ascending-ID selection biases toward earlier-digitised works; no genre or popularity filtering was applied.

Each text is chunked into passages of 50 tokens with 15-token overlap between consecutive chunks, producing 24,964,565 total passages. The 50-token window is short enough to capture a single narrative beat---a moment of dialogue, a description, a transition---while the 15-token overlap ensures that no content falls at a chunk boundary. Each passage is embedded using BGE-large-en-v1.5 \citep{xiao2024bge, baai2023bge} into a 1024-dimensional L2-normalised vector. All embeddings are precomputed.

\subsection{Temporal Co-occurrence Pairs}
\label{sec:pairs}

For each text, we extract all pairs of passages whose positions are within a window of 15 chunks. A passage at position $i$ is paired with every passage at positions $i{-}15$ through $i{+}15$ (excluding itself) within the same text. This produces 373,296,555 unique co-occurrence pairs across the corpus.

The 15-chunk window (approximately 525 tokens of non-overlapping content) captures the local narrative neighbourhood: passages that participate in the same scene, the same argument, the same descriptive movement. Passages 15 chunks apart are typically in the same narrative episode but may have shifted topic, tone, or speaker---the window is wide enough to link the setup of a scene with its resolution, but narrow enough to exclude unrelated narrative segments.

All pairs are within-book by construction---the pair generation script only creates pairs between passages within the same text. The corpus contains 8 cases of duplicate Gutenberg editions (the same work appearing under different Gutenberg IDs), affecting approximately 739,140 pairs (0.19\% of training data). These duplicates are retained rather than filtered, as they represent a negligible fraction of training signal. We verified that none of the five featured unseen evaluation novels have duplicate editions in the training corpus.

\subsection{Association Model}
\label{sec:model}

The model architecture follows AAR \citep{dury2026aar}: a 4-layer MLP with GELU activations, LayerNorm, and a learned residual connection:

\begin{equation}
\label{eq:association}
f(\mathbf{x}) = \mathrm{normalise}\bigl(\alpha \cdot \mathbf{x} + (1 - \alpha) \cdot g(\mathbf{x})\bigr)
\end{equation}

where $g$ is the MLP transformation, $\alpha$ is a learned scalar (converging to 0.756), and the output is L2-normalised. The hidden dimension matches the input (1024), yielding 29,404,161 parameters. The residual connection preserves the original embedding's semantic information while learning an associative perturbation---the model learns to \emph{adjust} the embedding geometry, not replace it.

Training uses symmetric contrastive loss (InfoNCE) with in-batch negatives: for a batch of 512 pairs, each positive pair is contrasted against 511 negatives. Temperature is fixed at $\tau = 0.05$. Training proceeds in two phases: 100 epochs of probe training followed by 50 epochs of warm-start training, for 150 total epochs. AdamW optimiser with cosine learning rate schedule.

The model reaches 42.75\% training accuracy at epoch 150 (loss 3.030). Training accuracy is the fraction of batches in which the correct positive is ranked first among all 512 in-batch candidates under the symmetric contrastive loss---effectively top-1 retrieval accuracy within each batch, averaged over both directions. This is well below the capacity ceiling for an architecture of this size. The gap between capacity and accuracy is consistent with a compression regime that forces generalisation beyond individual co-occurrence pairs (Section~\ref{sec:theory}).

\subsection{Clustering}
\label{sec:clustering}

We apply $k$-means clustering to the association-space embeddings (the output of $f$) at six granularities: $k = 50$, 100, 250, 500, 1,000, and 2,000. Clustering is performed over all 24.96 million passages. Each passage receives a cluster assignment at every $k$ value, enabling multi-resolution analysis.

For each $k$ value, we apply quality filters based on minimum book diversity thresholds to remove degenerate clusters. Passing clusters and summary statistics at each level (Table~\ref{tab:cluster-stats}):

\begin{table}[t]
\centering
\caption{Cluster statistics at each granularity. ``Mean Books'' is the average number of distinct books per cluster. ``Mean Dominance'' is the average maximum single-book fraction. ``Mean Cosine'' is mean intra-cluster cosine similarity in association space.}
\label{tab:cluster-stats}
\begin{tabular}{@{}cccccc@{}}
\toprule
$k$ & Threshold & Passing & Mean Cosine & Mean Books & Mean Dominance \\
\midrule
50 & 50 & 50/50 & 0.302 & 5,860 & 2.2\% \\
100 & 100 & 100/100 & 0.378 & 4,508 & 4.0\% \\
250 & 200 & 241/250 & 0.436 & 3,329 & 6.3\% \\
500 & 500 & 472/500 & 0.479 & 2,502 & 9.0\% \\
1,000 & 500 & 857/1,000 & 0.513 & 1,885 & 12.0\% \\
2,000 & 1,000 & 980/2,000 & 0.499 & 1,797 & 8.1\% \\
\bottomrule
\end{tabular}
\end{table}

At $k = 100$, each cluster draws from an average of 4,508 books---nearly half the corpus---establishing that the discovered patterns are cross-author, not idiosyncratic. Mean dominance of 4.0\% confirms no cluster is driven by a single text. The pattern across $k$ values is consistent: as granularity increases, clusters become tighter (higher cosine), more specific (fewer books), and more susceptible to single-book dominance---all expected consequences of finer partitioning.

\subsection{From Episodic Recall to Concept Formation}
\label{sec:theory}

The theoretical relationship between PAM and concept discovery requires explanation. PAM \citep{dury2026pam} was designed for single-pass, sequential experience. An agent experiences a stream of states, and a predictor that looks across the full timeline of stored experience learns which past states are associatively reachable from the current state. The goal is faithful episodic recall---remembering what was experienced, from the perspective at which it was experienced. In that regime, memorisation is correct behaviour and inductive transfer is expected to fail (associations are specific to experienced co-occurrences).

The corpus method operates in a different regime. The text is fixed and replayable. Multi-epoch training replays the same temporal co-occurrences repeatedly, analogous to hippocampal sleep replay consolidating episodes into stable neocortical representations \citep{mcclelland1995cls, wilson1994reactivation}. The capacity bottleneck---29.4M parameters attempting to encode 373 million co-occurrence relationships---makes full memorisation of individual co-occurrence relations infeasible. Instead, the model must find regularities: patterns of transition structure that recur across many books.

Consider what happens when a ``moment of departure'' appears in a Jane Austen novel, a Western, a Gothic horror, and a philosophical essay. Each instance creates co-occurrence pairs linking the departure passage to its specific neighbours. Across hundreds of such instances, the model cannot store each individual mapping. What it can do is learn that certain passages---regardless of their semantic content---occupy a similar position in transition space. They have similar predecessors (building tension, stating stakes) and similar successors (journey, new setting, uncertainty). This shared relational signature is what we call a \emph{concept}---a recurrent pattern of transition structure. ``Narrative function'' is the most intuitive label for many such concepts (confrontation, departure, revelation), but the method also discovers discourse registers, literary traditions, and scene templates through the same mechanism.

This is how PAM's episodic recall mechanism becomes concept formation: same training signal (temporal co-occurrence), different compression regime. Under severe capacity constraint, specific associations get compressed into structural patterns. PAM provides the principle; contrastive replay under bottleneck provides the consolidation.

This also explains the divergence in inductive transfer. In AAR \citep{dury2026aar}, passage-to-passage associations are contingent---a passage about Quentin Tarantino and a passage about Knoxville, Tennessee are associated only because they happen to answer the same question. There is no structural regularity that would let a model predict this association from unseen pairs. In the corpus setting, transition-structure patterns are recurrent---the ``moment of departure'' appears thousands of times across thousands of books. A model that has learned this structural pattern can recognise it in an unseen novel, because the pattern is a recurrent regularity in literary transition structure, not a contingent co-occurrence.

\section{Experimental Setup}
\label{sec:setup}

\subsection{Corpus}
\label{sec:corpus}

The training corpus comprises 9,766 English-language texts from Project Gutenberg (10,000 requested, 234 excluded during chunking). Texts span approximately four centuries of English-language writing, including fiction, essays, drama, poetry, religious texts, biography, and history. No genre labels or metadata beyond author and title are used in training. The corpus produces 24,964,565 passages and 373,296,555 co-occurrence pairs after chunking.

\subsection{Baselines}
\label{sec:baselines}

\textbf{Raw BGE clustering.} We cluster passages using their original BGE-large-en-v1.5 embeddings (before association-space transformation) at $k = 100$ and $k = 250$. This provides a direct comparison: same passages, same clustering algorithm, but organised by semantic similarity instead of learned association. Due to computational constraints, BGE clustering is computed on a 2,000-novel subset of the corpus (8,153,406 passages).

For a matched comparison on the same 2K corpus, we also ran PAM at $k = 100$:

\begin{table}[t]
\centering
\caption{Matched 2K comparison. The PAM 10K column is included for cross-scale reference, not as a matched comparison.}
\label{tab:matched-2k}
\begin{tabular}{@{}lccc@{}}
\toprule
Metric & PAM 2K ($k{=}100$) & BGE 2K ($k{=}100$) & PAM 10K ($k{=}100$) \\
\midrule
Mean cosine & 0.454 & 0.493 & 0.378 \\
Mean books & 1,121 & 1,905 & 4,508 \\
Mean dominance & 7.9\% & 2.3\% & 4.0\% \\
\bottomrule
\end{tabular}
\end{table}

At 2K scale, PAM clusters are tighter and more concentrated than BGE---fewer books per cluster, higher dominance. At 10K scale, the pattern reverses: PAM clusters become far more diverse (4,508 books versus BGE's 1,905). This scale-dependent behaviour is consistent with the compression hypothesis (Section~\ref{sec:theory}): at 2K, the model has sufficient capacity relative to data to learn author-specific and tradition-specific patterns; at 10K, the capacity bottleneck forces cross-author generalisation, producing clusters that span many more books. The qualitative character of the clusters also differs: 2K PAM clusters tend toward author-tradition groupings, while 10K PAM clusters capture broader functional modes.

The higher cosine in BGE clusters reflects the same underlying distinction at both scales: similarity-based clusters are semantically tight (high cosine) because they select for shared vocabulary, while association-based clusters are semantically broader because they select for shared transition structure, which crosses vocabulary boundaries.

\textbf{Context-enriched baseline.} To test whether PAM's learned transformation provides signal beyond simple local context, we compute a non-learned context-enriched embedding for each passage by averaging the BGE embeddings within the same $\pm$15-chunk window used for PAM training pairs (respecting book boundaries), then L2-normalising and clustering. This tests whether the structure PAM discovers could be recovered by a symmetric averaging operation without a learned contrastive transformation.

\begin{table}[t]
\centering
\caption{Context-enriched baseline vs BGE and PAM on the matched 2K corpus.}
\label{tab:context-enriched}
\begin{tabular}{@{}lccc@{}}
\toprule
Metric ($k{=}100$) & BGE Raw & Context-Enriched & PAM 2K \\
\midrule
Passing & 100/100 & 95/100 & 100/100 \\
Mean cosine & 0.493 & 0.861 & 0.454 \\
Mean books & 1,905 & 726 & 1,121 \\
Mean dominance & 2.3\% & 8.1\% & 7.9\% \\
\bottomrule
\end{tabular}
\end{table}

Context averaging massively inflates intra-cluster cosine ($0.493 \rightarrow 0.861$)---smoothing makes nearby passages look similar, but this likely reflects local continuity and within-book proximity rather than the broader cross-book structural regularities of interest here. More critically, it collapses book diversity ($1{,}905 \rightarrow 726$) and increases dominance ($2.3\% \rightarrow 8.1\%$), producing book-specific clusters rather than cross-book concepts. At $k = 250$, the pattern worsens: context-enriched clustering fails 28/250 quality filters (versus 9/250 for PAM 10K), with mean books dropping to 465 and dominance reaching 12.9\%.

PAM is substantially broader than context-averaging (1,121 vs 726 books per cluster) while maintaining much lower cosine (0.454 vs 0.861)---PAM's clusters are semantically diverse rather than locally smoothed. The learned contrastive transformation extracts cross-book structural patterns that symmetric context averaging cannot recover. Context averaging captures passages that are similar because they are nearby in the same book; PAM better preserves cross-book structural regularities that simple local averaging cannot recover.

\textbf{Random MLP baseline.} We apply $k = 100$ clustering to the output of a randomly initialised MLP with the same architecture but no training. This tests whether the architecture itself (the residual connection, the normalisation) imposes structure that could be mistaken for learned concepts. The random MLP produces 99/100 passing clusters with mean cosine 0.473, mean book diversity 8,553, and mean dominance 1.1\%. The near-uniform book distribution (8,553 of 9,766 books per cluster) confirms that the random MLP imposes no meaningful structure---it disperses passages nearly uniformly. PAM's selectivity (4,508 books per cluster versus 8,553) is a consequence of learned structure, not architectural bias.

\subsection{Validation Controls}
\label{sec:controls}

\textbf{Position-in-book.} For each $k = 100$ cluster, we compute the mean normalised position (0 = start, 1 = end) of all passages. Clusters where the mean falls outside [0.3, 0.7] or the standard deviation is below 0.15 are flagged. Result: 0/100 clusters flagged. Mean positions cluster tightly around 0.50, indicating no systematic position bias.

\textbf{Token count.} Clusters where the mean token count deviates by more than 3 standard deviations from the corpus mean are flagged. Result: 2/100 clusters flagged (clusters 34 and 84)---edge cases rather than systematic confounds.

\textbf{Book concentration.} Clusters where any single book contributes more than 10\% of passages are flagged. Result: 10/100 clusters exceed 10\% dominance. The most extreme is a German-language cluster (cluster 34, 15.0\% dominance, only 292 books)---an expected artefact in a predominantly English corpus. The remaining flagged clusters are driven by very long books that naturally contribute many passages to functionally coherent clusters. All other flagged clusters still contain more than 1,300 distinct books.

\textbf{Temporal shuffle control.} On a 2,000-novel pilot corpus trained with the same architecture, randomly permuting temporal ordering within each text (preserving all passage embeddings but destroying co-occurrence structure) collapses cross-boundary recall by 95.2\%. This confirms the model learns genuine temporal co-occurrence structure, not artefacts of embedding geometry. The shuffle control was performed on the pilot corpus (which reached 51.0\% accuracy at 100 epochs), not the full 10K corpus; the same architecture and training procedure were used in both.

\subsection{Unseen-Novel Evaluation}
\label{sec:unseen-setup}

Five canonical novels absent from the training corpus were processed through the trained model for inductive evaluation. These were not formally held out from a sampling frame---they are simply well-known texts that do not appear among the 9,766 training texts. We confirmed the absence of each from the training corpus by Gutenberg ID and verified that no duplicate editions exist in the training set.

\begin{table}[t]
\centering
\caption{Unseen evaluation novels.}
\label{tab:unseen-novels}
\begin{tabular}{@{}llll@{}}
\toprule
Novel & Gutenberg ID & Author & Chunks \\
\midrule
Pride and Prejudice & 1342 & Jane Austen & 4,697 \\
Dracula & 345 & Bram Stoker & 5,879 \\
Frankenstein & 84 & Mary Shelley & 2,638 \\
Alice's Adventures in Wonderland & 11 & Lewis Carroll & 1,057 \\
The War of the Worlds & 36 & H.G.\ Wells & 2,157 \\
\bottomrule
\end{tabular}
\end{table}

Each novel's passages are embedded with BGE-large-en-v1.5 and transformed through the trained association model, then assigned to the nearest existing cluster centroid at each $k$ value. No retraining or fine-tuning occurs. For comparison, the same novels are also assigned to centroids from raw BGE clustering on the 2,000-novel subset (Section~\ref{sec:baselines}); the scale mismatch (2K BGE vs 10K PAM centroids) is a known limitation. Five additional novels (Sherlock Holmes, A Tale of Two Cities, Jane Eyre, The Picture of Dorian Gray, Moby Dick) are available for inspection in the interactive demonstration. Cluster labels were generated from training-corpus passages (Section~\ref{sec:clustering}); unseen novels were assigned to pre-labelled clusters, so the labels are independent of the evaluation texts.

\section{Results}
\label{sec:results}

\subsection{Association Space Discovers Transition Structure}
\label{sec:transition}

\begin{figure}[!htbp]
\centering
\includegraphics[width=\textwidth]{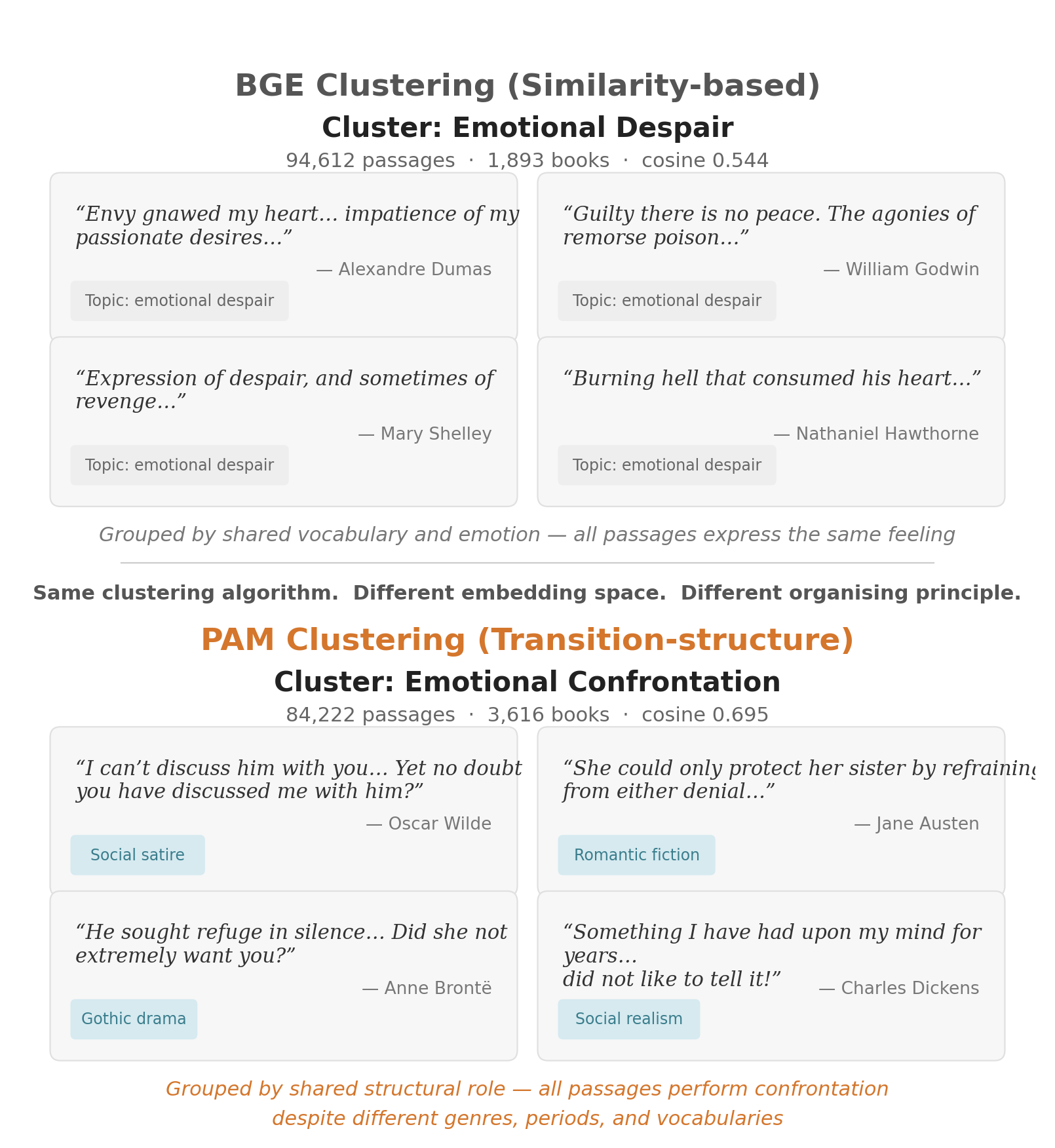}
\caption{BGE similarity-based clusters group passages by topic (what text is \emph{about}), while PAM association-based clusters group by transition structure (what text \emph{does})---narrative function, discourse register, and literary tradition.}
\label{fig:bge-vs-pam}
\end{figure}

The clearest way to see the difference is to compare clusters directly: association-space clusters group passages by transition structure---narrative function, discourse register, literary tradition---while embedding-similarity clusters group by topic (Figure~\ref{fig:bge-vs-pam}).

\textbf{BGE clusters (similarity-based)} organise by content. Representative clusters from the 2K baseline include groupings of all passages mentioning financial transactions, all passages describing fear or dread, all passages with nautical vocabulary, and all passages with religious language. A passage describing fear during a chase scene and a passage describing fear during a quiet domestic moment occupy the same BGE cluster---they are \emph{about} the same thing.

\textbf{PAM clusters (association-based)} organise by transition structure. Representative clusters at $k = 100$ include:

\textbf{``Direct confrontation and negotiation''} (460,753 passages, 5,088 books---52\% of all books in the corpus). Scenes where characters face each other with competing demands---spanning diplomatic negotiations in historical fiction, drawing-room arguments in domestic novels, interrogation scenes in mysteries, and power struggles in adventure fiction. The passages share no vocabulary; what they share is the narrative beat of two parties stating positions.

\textbf{``Cynical worldly wisdom''} (394,317 passages, 5,202 books---53\% of books). Passages delivering hard-won pragmatic observations about human nature, found across 18th-century satire, Victorian social commentary, American realism, and philosophical essays. The functional signature is a voice standing slightly outside the action and commenting on it.

\textbf{``Lyrical landscape meditation''} (368,654 passages, 5,924 books---61\% of books). Extended descriptive passages where prose rhythm slows and sensory detail accumulates---spanning Romantic nature writing, Gothic atmosphere-setting, travel writing, and pastoral fiction. The function---slowing the reader, establishing mood through accumulation---is shared across genres that have nothing else in common.

\textbf{``Detective investigation and inquiry''}. Not limited to detective fiction. This cluster captures the narrative mode of systematic questioning and evidence-gathering wherever it appears---in mystery novels, legal dramas, journalistic expos\'{e}s, and Gothic investigations of the supernatural. In the interactive demonstration, a passage from Dr Jekyll and Mr Hyde---where Mr Utterson begins asking about a mysterious door---is assigned to this cluster despite containing no detective vocabulary. The passage has the investigative \emph{frame}: cautious questioning, guarded responses, the promise of a strange story to come.

\FloatBarrier

\textbf{``Discovery of death or horror''} (99,642 passages at $k = 250$, 3,728 books). The cluster from which our introductory example is drawn. It contains passages from H.\ Rider Haggard's \emph{She} (a South American exploration), Algernon Blackwood's \emph{The Damned} (an English country house), H.G.\ Wells's science fiction, Jeffery Farnol's romance, an Australian bush memoir, and Gothic horror---unified by the narrative beat of a character alone, encountering something wrong, and the body responding before the mind catches up.

The book diversity numbers depend on scale. On the matched 2K corpus (Table~\ref{tab:matched-2k}), PAM clusters are actually more concentrated than BGE---1,121 books per cluster versus 1,905. The qualitative difference is already present (PAM groups by transition structure, BGE by topic), but the clusters tend toward author-tradition groupings. At 10K scale, the pattern flips: PAM clusters average 4,508 books (46\% of the corpus), spanning traditions, centuries, and genres. We attribute this to compression pressure---the same architecture forced to fit $4.5\times$ more data extracts broader patterns. The claim is not that PAM is uniformly broader than BGE, but that increasing compression shifts the method from tradition-linked structure toward cross-author concepts.

\subsection{Multi-Resolution Concept Structure}
\label{sec:multiresolution}

\begin{figure}[!htbp]
\centering
\includegraphics[width=\textwidth]{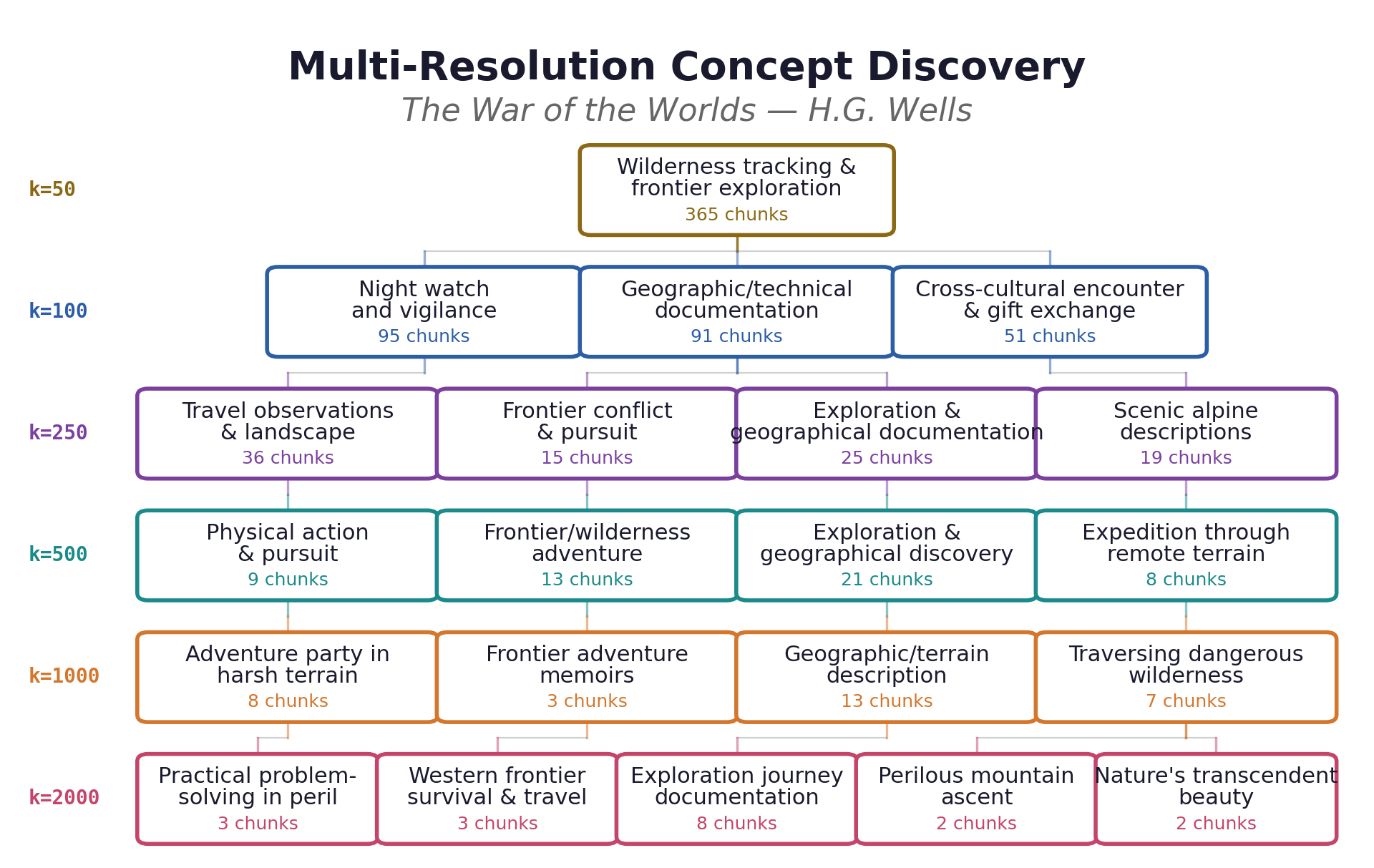}
\caption{Multi-resolution concept structure from $k{=}50$ to $k{=}2{,}000$. Broad narrative modes at coarse resolution decompose into increasingly specific variants---registers, traditions, and scene templates---at finer granularity.}
\label{fig:multiresolution}
\end{figure}

The six $k$ values reveal structure that is hierarchical on inspection: broad modes at $k = 50$ decompose into increasingly specific variants at higher resolution (Figure~\ref{fig:multiresolution}).

\FloatBarrier

\textbf{Historical conflict lineage.} At $k = 50$, a broad ``historical combat and retreat'' cluster captures all passages describing military engagements. At $k = 100$, this separates into distinct modes including battlefield narrative and political manoeuvring. By $k = 1{,}000$, recognisable subcategories emerge: Ottoman/Crusader wars, Napoleonic campaigns, American Civil War. At $k = 2{,}000$, these sharpen to highly specific clusters: Franco-Prussian War 1870--71, specific Napoleonic battle accounts, American Revolutionary War correspondence.

\textbf{Legal/judicial lineage.} At $k = 50$, a broad ``legal proceedings and authority'' cluster encompasses all passages involving law, judgment, and institutional power. At $k = 100$, this separates into distinct modes including legal/criminal case discussion (195,082 passages, 6,078 books) and related clusters for interrogation and crime-and-consequence. By $k = 500$, recognisable sub-modes emerge: courtroom proceedings, interrogation scenes, and crime narratives. At $k = 1{,}000$, these resolve further: legal testimony, evidence presentation, and---as a distinct branch---witchcraft trials, where the legal frame meets the supernatural subject matter. At $k = 2{,}000$, courtroom testimony and witch trial proceedings are separate, highly specific clusters.

\textbf{Emergent categories.} Several $k = 1{,}000$ and $k = 2{,}000$ clusters defy conventional genre or topic categories, representing functional patterns with no standard literary-critical label:

\begin{itemize}
    \item \textbf{Music performance scenes} ($k = 1{,}000$): Passages describing the act of playing or listening to music---concert halls, parlour performances, church organs, street musicians---united by prose attempting to render auditory experience in text.
    \item \textbf{Witchcraft and folk magic} ($k = 1{,}000$): Passages from folklore collections, anthropological accounts, historical trials, and fiction, united by the explanatory frame of sympathetic magic.
    \item \textbf{Sailor dialect} ($k = 1{,}000$): Passages using maritime vernacular across adventure fiction, naval history, and comic sketches---prose performing a specific sociolect.
    \item \textbf{Cats} ($k = 1{,}000$): Passages about cats (not animals generally) drawn from children's literature, domestic fiction, natural history, and essays---a specific prose register that is affectionate, observational, and slightly anthropomorphising.
    \item \textbf{Darwin-Huxley correspondence} ($k = 2{,}000$): The epistolary conventions of Victorian scientific correspondence---a register so specific that the model isolates it as a distinct functional pattern.
    \item \textbf{Social contract philosophy} ($k = 2{,}000$): Discursive passages in the style of Enlightenment political philosophy---Locke, Rousseau, and their inheritors---a specific argumentative register crossing centuries.
\end{itemize}

The discovered categories extend beyond any single literary framework. The method is not constrained to pre-defined taxonomies---it finds whatever transition patterns recur with sufficient regularity.

\begin{figure}[!htbp]
\centering
\includegraphics[width=\textwidth]{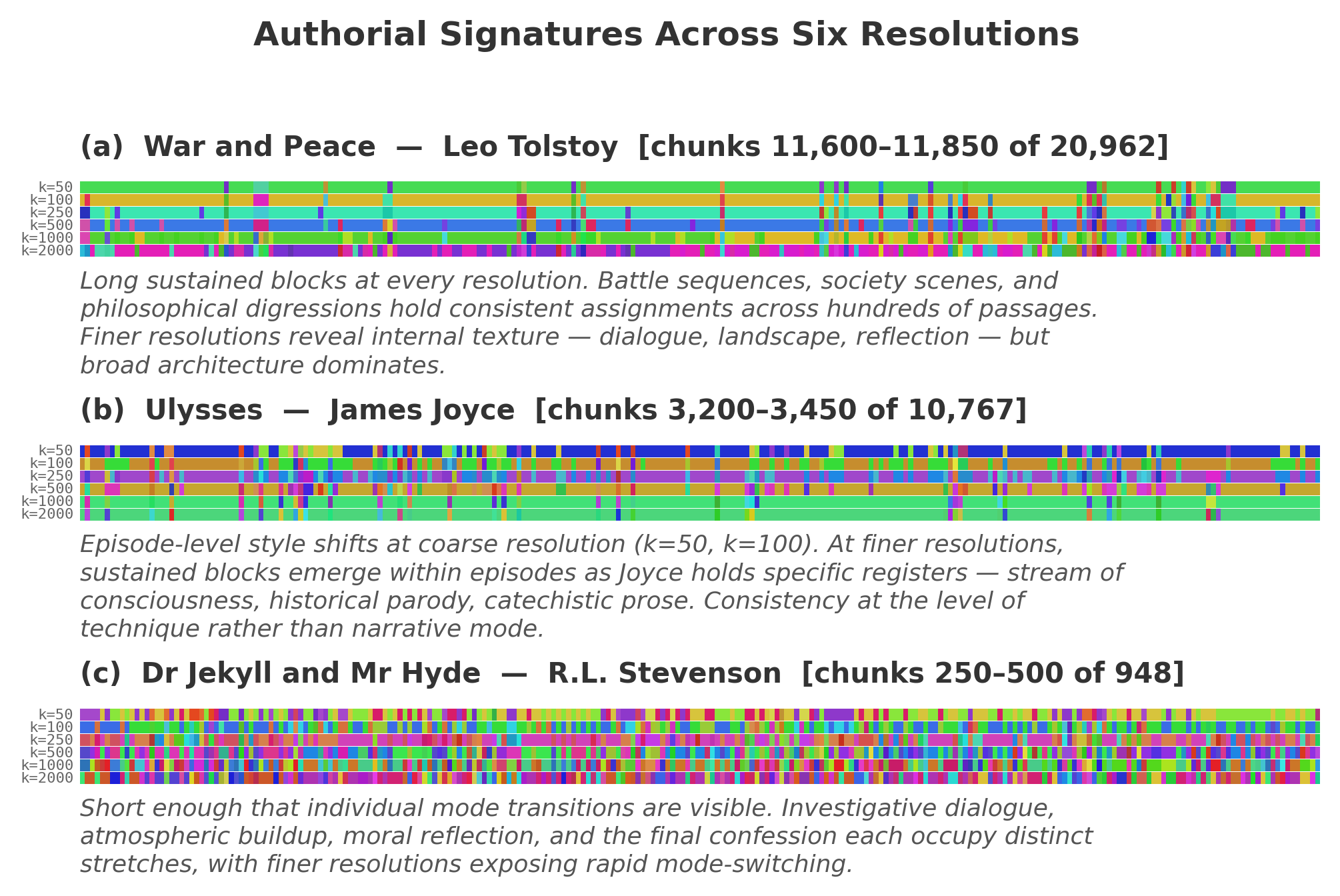}
\caption{Authorial pacing signatures for three canonical works. Different authors produce characteristically different distributions of structural modes, visible at multiple resolutions simultaneously.}
\label{fig:signatures}
\end{figure}

\textbf{Authorial pacing signatures.} The multi-resolution view also reveals how individual authors distribute structural weight across a text---a property we term \emph{pacing signature} (Figure~\ref{fig:signatures}).

Comparing three canonical works illustrates the range of strategies the concept map captures.

War and Peace (Tolstoy, 20,962 chunks) is structured in long, sustained blocks visible at every resolution. Extended battle sequences, society scenes, and philosophical digressions maintain consistent cluster assignments across hundreds of passages. At finer resolutions, internal texture emerges within these blocks---dialogue, landscape, and reflection interwoven---but the broad architecture remains dominant.

Ulysses (Joyce, 10,767 chunks) shows the inverse pattern. At coarse resolution, the novel's famous episode-level style shifts produce dramatic variation---each episode employs a distinct mode. At $k{=}1000$ and $k{=}2000$, sustained blocks emerge \emph{within} episodes as Joyce holds a specific register---stream of consciousness, historical parody, catechistic prose---for extended stretches. Consistency appears at the level of technique rather than broad narrative mode.

Dr Jekyll and Mr Hyde (Stevenson, 948 chunks) is short enough that individual mode transitions are visible at every resolution. The concept map reveals how a tightly constructed Gothic mystery distributes its structural weight: investigative dialogue, atmospheric buildup, moral reflection, and the final confession each occupy distinct stretches, with finer resolutions exposing rapid mode-switching within scenes.

These pacing signatures emerge from temporal co-occurrence statistics alone---no genre labels, no structural annotations, no author metadata. They suggest a potential lens for comparative literary analysis: different authors appear to produce characteristically different distributions of structural modes, and these distributions are visible at multiple resolutions simultaneously. We present them as an exploratory application of the concept map rather than a validated analytical framework.

\subsection{Cross-Genre Evidence}
\label{sec:crossgenre}

The cross-genre examples (identified by systematic search across cluster readouts) provide the strongest evidence that clustering reflects structure, not topic. Genre categories were assigned post-hoc from author and title metadata for this analysis only---no genre labels were used in training.

\textbf{Formal confrontation/negotiation} ($k = 250$, Cluster 85, 111,336 passages, 3,675 books, 6 genre categories). This cluster contains passages from George MacDonald's religious fiction, Wilkie Collins's sensation novels, a Shaw-influenced socialist satire, historical military fiction, Gothic mystery, and domestic romance. In every case, the narrative beat is identical: two characters in controlled adversarial dialogue, stating demands, revealing knowledge, establishing power dynamics. The vocabulary, setting, period, and genre differ entirely.

\textbf{Dramatic confrontation and warning} ($k = 500$, Cluster 254, 61,098 passages, 3,806 books, 5 genres). Passages from Lewis's \emph{The Monk} (Gothic horror), Hugo's \emph{Notre-Dame de Paris} (historical romance), Marie Corelli (sensation fiction), Frederick the Great historical fiction, and satirical fiction---all sharing the operatic register of ``you have destroyed me'' declarations. The function is the same: a character at extremity, making a speech of accusation or prophecy.

\textbf{Deathbed/medical crisis} ($k = 500$, Cluster 157, 61,443 passages, 4,157 books, 5 genres). Passages from adventure fiction, philosophy essays, religious devotional writing, domestic romance, and Mark Twain's satire---all describing a physician attending a patient in crisis, observers watching anxiously, and the clinical vocabulary of decline. The scene template---medical authority confronting human fragility while loved ones wait---recurs across every genre that includes mortality.

\subsection{Inductive Transfer to Unseen Novels}
\label{sec:inductive}

Five canonical novels absent from the training corpus demonstrate that the discovered concepts generalise to unseen texts.

\begin{figure}[!htbp]
\centering
\includegraphics[width=\textwidth]{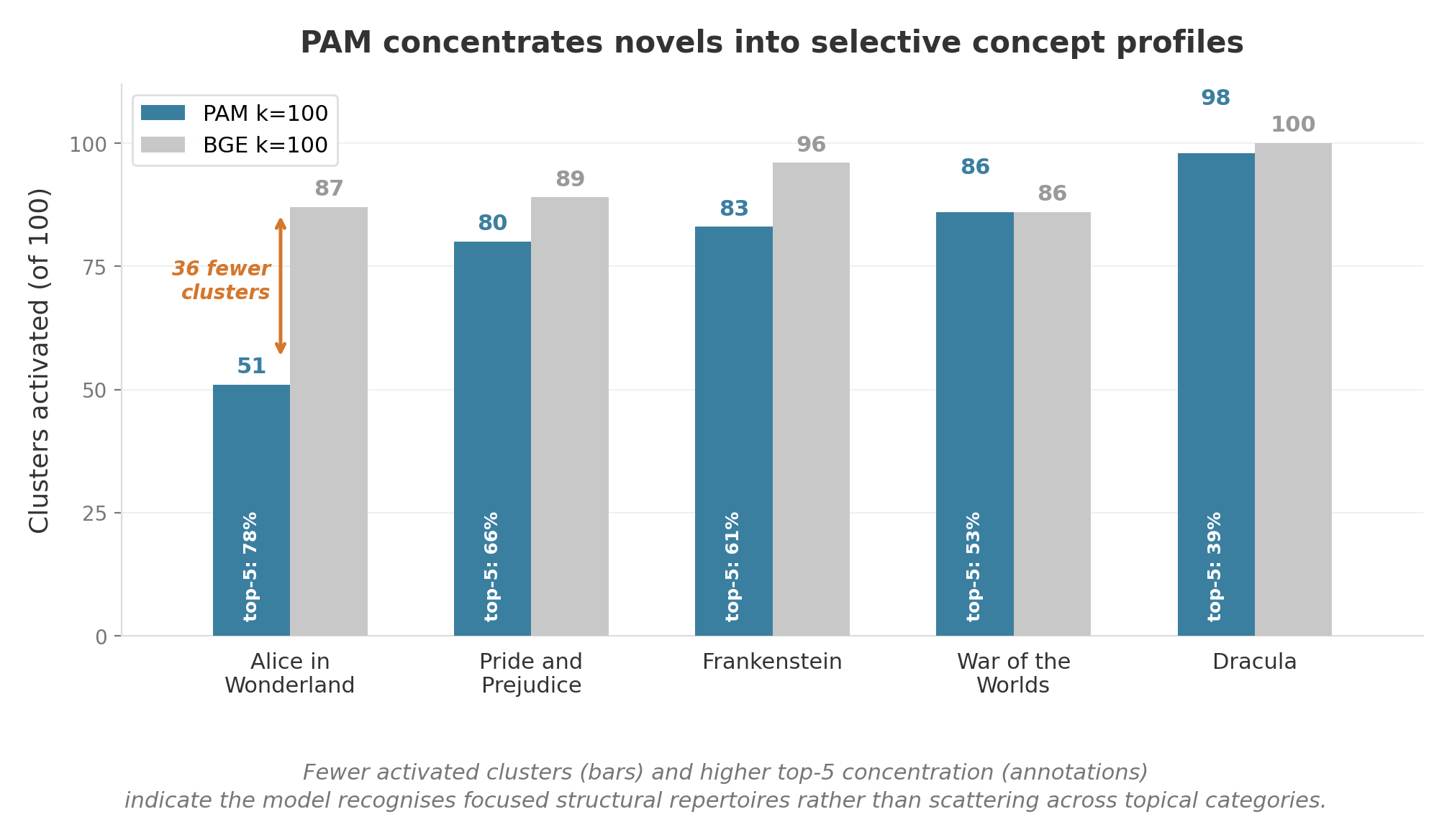}
\caption{Cluster selectivity for unseen novels. PAM concentrates each novel into a selective subset of structurally coherent clusters, while BGE assignment saturates nearly all clusters.}
\label{fig:selectivity}
\end{figure}

\textbf{Selectivity.} The association model concentrates each novel into a selective subset of clusters; raw BGE assignment spreads each novel across nearly all clusters (Table~\ref{tab:selectivity}; Figure~\ref{fig:selectivity}).

\begin{table}[t]
\centering
\caption{Cluster activation and top-5 concentration for unseen novels, ordered by PAM selectivity. ``Top-5 PAM/BGE'' is the fraction of a novel's passages assigned to its five most frequent clusters---higher means a more concentrated concept profile.}
\label{tab:selectivity}
\begin{tabular}{@{}lcccc@{}}
\toprule
Novel & PAM $k{=}100$ & BGE $k{=}100$ & Top-5 PAM & Top-5 BGE \\
\midrule
Alice in Wonderland & 51/100 & 87/100 & 77.6\% & 32.2\% \\
Pride and Prejudice & 80/100 & 89/100 & 66.5\% & 25.2\% \\
Frankenstein & 83/100 & 96/100 & 60.6\% & 42.4\% \\
The War of the Worlds & 86/100 & 86/100 & 52.6\% & 36.7\% \\
Dracula & 98/100 & 100/100 & 39.1\% & 19.5\% \\
\bottomrule
\end{tabular}
\end{table}

Alice in Wonderland is the most striking case. PAM assigns its passages to only 51 of 100 clusters, while BGE assigns to 87. More than three-quarters of Alice's passages (77.6\%) fall into just five PAM clusters, dominated by two: ``Domestic ritual and children's play'' (34.5\%) and ``Domestic incident recounting'' (20.5\%). Carroll's novel touches many topics---tea parties, trials, gardens, croquet, playing cards---but employs a remarkably narrow range of structural modes. The Mad Hatter's tea party and the Queen's croquet game are semantically different; PAM recognises them as structurally identical (both assigned to the same ``games and rituals with arbitrary rules'' cluster). BGE, sorting by vocabulary, treats them as unrelated.

The \emph{absence} of certain clusters is equally telling. ``Solitary journey with introspection''---which dominates both Frankenstein (21.9\%) and The War of the Worlds (24.1\%)---appears in only 8 of Alice's 1,057 passages (0.8\%). Alice doesn't introspect; she reacts. The cluster assignments pick up on this difference without being told anything about the novels.

Pride and Prejudice tells a complementary story. PAM's top cluster---``Romantic entanglements and gossip''---accounts for 29.7\% of the entire novel. Nearly a third of Austen's text, in PAM's reading, is performing a single structural mode: characters circling questions of attachment, propriety, and reputation. BGE would group these as ``passages about romance.'' PAM groups them as a specific \emph{mode of discourse}---speculative, socially attentive, status-conscious---that Austen sustains across scenes about dances, letters, walks, and drawing-room conversations. BGE's top-5 hold only 25.2\%; it sees Austen's vocabulary diversity, not her structural consistency.

Dracula sits at the opposite end: its top-5 PAM clusters hold only 39.1\%---a flatter distribution than any other test novel. That fits: Dracula is a multi-genre novel---travel journal, personal diary, newspaper clipping, ship's log, Gothic horror, medical case notes. The lower concentration reflects a genuinely varied repertoire.

Frankenstein activates 83 PAM clusters but concentrates heavily: its top-5 hold 60.6\% of passages. Two modes dominate---solitary introspection (21.9\%) and emotional confrontation (13.2\%)---capturing the novel's dual obsession: Victor's guilt-driven isolation and the Creature's abandoned rage. ``Scholarly or artistic reflection'' ranks third at 11.4\%, grouping Victor's natural philosophy and the Creature's self-education---two parallel intellectual journeys that PAM assigns to the same concept despite different narrators and vocabularies.

The War of the Worlds activates 86 PAM clusters---coincidentally matching BGE's count---but with much higher concentration (top-5: 52.6\% vs 36.7\%). The novel's two-part structure is visible in the assignments: Book 1 alternates between solitary observation (24.1\%), tense survivor dialogue (12.3\%), and military engagement (6.7\%); Book 2 opens with a distinct travel-adventure register as refugees flee down the Thames. ``Night watch and vigilance'' (4.4\%) maps onto exactly the atmospheric scenes that define Wells's technique---the narrator lying awake listening for tripod footsteps, watching heat-ray flashes on the horizon. A cluster trained on 9,766 texts applies cleanly to an unseen novel.

At $k = 2{,}000$, the selectivity pattern sharpens further: Pride and Prejudice activates only 27.5\% of PAM clusters; Dracula activates 48.8\%.

\textbf{Tracking structural role.} PAM assignments track structural role and shift at mode boundaries, while BGE assignments track topic and scatter with every vocabulary change.

In Dracula, the travel journal sections maintain consistent PAM cluster assignment through changes in described scenery---the textual mode (observational travel writing) is stable even as the topic shifts from mountains to villages to food. When the mode changes---from travel observation to supernatural dread---the PAM assignment changes correspondingly, even when surface vocabulary (descriptions of a castle) remains similar. BGE labels shift with every new described object.

In Frankenstein, the Creature's monologue (beginning at approximately 42\% of the novel) triggers a sustained shift to the ``Solitary journey with introspection'' cluster that holds through the De Lacey family episodes---a single structural mode covering the novel's longest unbroken stretch of isolated first-person reflection. At the trial of Justine ($\sim$32\%), PAM briefly surfaces ``Legal/criminal case discussion'' and ``Interpretive uncertainty,'' correctly identifying the judicial framing as structurally distinct from the surrounding emotional register.

In Pride and Prejudice, dialogue scenes cluster together under PAM regardless of whether the topic is marriage proposals, social slights, or family finances. The structural mode---witty adversarial exchange---is the stable signal. BGE separates these scenes by what is being discussed.

\subsection{Compression and Concept Formation}
\label{sec:compression}

The relationship between training accuracy and concept quality provides indirect evidence for the compression hypothesis.

The 2K pilot (2,000 novels, 8.15 million passages) reached 51.0\% accuracy at 100 epochs with the same architecture. The 10K corpus (9,766 novels, 24.96 million passages) reaches 42.75\% at 150 epochs. The model's capacity is identical; the lower accuracy reflects the increased compression pressure of fitting $4.5\times$ more co-occurrence relationships into the same parameter budget.

Both accuracy levels produce coherent, functionally meaningful clusters. The 10K clusters are qualitatively richer---more specific at fine $k$ values, more diverse in book coverage---consistent with the model having more varied examples of each narrative pattern to compress.

We note an important caveat: we have not performed a systematic accuracy sweep to identify an optimal compression ratio. The 42.75\% accuracy is a single data point from a single training run, not a validated sweet spot. What we can say is that at this compression level, the model produces clusters that are demonstrably functional rather than topical, and that the higher compression at 10K scale did not degrade cluster quality relative to the 2K pilot.

\section{Discussion}
\label{sec:discussion}

\subsection{What These Concepts Are}
\label{sec:what}

A passage's concept assignment, as discovered by this method, reflects its \emph{relational signature in transition space}. Two passages belong to the same concept if they tend to be preceded by similar kinds of passages and followed by similar kinds of passages, across many different texts. The method does not encode what a passage says; it encodes where that passage sits in the local structure of a text.

On inspection, the discovered concepts span at least five recognisable categories that literary analysis would normally treat separately:

\begin{itemize}
    \item \textbf{Narrative functions}---confrontation, investigation, departure, revelation---correspond to Propp's morphological functions and similar frameworks, but discovered empirically from transition statistics rather than imposed by analysts.
    \item \textbf{Discourse registers}---cynical worldly wisdom, lyrical landscape meditation, sailor dialect---capture modes of address or prose style that recur across genres.
    \item \textbf{Literary traditions}---Russian psychological realism, American Western, epistolary prose---capture author-community conventions that produce similar transition patterns.
    \item \textbf{Scene templates}---deathbed/medical crisis, domestic interruption, formal negotiation---capture recurring situational structures.
    \item \textbf{Subject-matter conventions}---cats, witchcraft, Darwin-Huxley correspondence---capture recurring treatments of specific subject matter where the treatment (not merely the topic) is the shared signal.
\end{itemize}

All of these are organised by transition structure, not semantic content. A topic model would group all cat passages together regardless of register; our method separates the affectionate-observational cat register from, say, zoological description of felids, because these occupy different positions in transition space. Conversely, our method would be expected to group a sardonic aside about cats with a sardonic aside about dogs, because the discourse register (not the subject) determines the transition neighbourhood.

We use ``concept'' as the umbrella term for these recurrent transition-structure patterns. At coarse $k$ values, most concepts correspond to recognisable narrative or discourse modes. At fine $k$ values, they resolve into specific traditions, registers, and conventions---the actual structure of English-language literary production, not any pre-specified categorisation.

\subsection{Relationship to PAM and AAR}
\label{sec:relationship}

The three papers in this line of work---PAM, AAR, and the present paper---share a training signal (temporal co-occurrence) and a training objective (contrastive learning over embeddings) but operate in different regimes with different emergent properties:

\begin{table}[t]
\centering
\caption{Comparison of PAM, AAR, and this work across key properties.}
\label{tab:comparison}
\begin{tabular}{@{}llll@{}}
\toprule
Property & PAM & AAR & This Work \\
\midrule
Domain & Synthetic benchmark & Multi-hop QA & Literary corpus \\
Goal & Faithful episodic recall & Passage retrieval & Concept discovery \\
Compression & None (memorisation) & Minimal ($\sim$97\%) & Severe (42.75\%) \\
Inductive transfer & Fails & Fails & Succeeds \\
What is learned & Specific associations & Corpus-specific links & Structural patterns \\
Biological analogy & Hippocampal encoding & --- & Neocortical consolidation \\
\bottomrule
\end{tabular}
\end{table}

The results point to compression as the key variable. PAM memorises because it should---episodic recall requires remembering specific events. AAR reaches $\sim$97\% training accuracy, which means it too learns specific links. The present work reaches 42.75\%, which forces something different: pattern extraction. The same architecture and training signal produce different behaviour under different compression regimes.

The inductive transfer results align with this interpretation. AAR's failure to transfer (+0.10 on HotpotQA, $-$7.6 on MuSiQue) reflects the specificity of learned associations---each is contingent on a particular corpus. Preliminary cross-domain results suggest partial transfer in domains with conserved physical constraints, where associations reflect structural regularities rather than contingent co-occurrences. The present work's successful transfer to unseen novels reflects the full regularity of narrative patterns---they are structural features of literary transition patterns, not contingent properties of specific texts.

\subsection{Practical Applications}
\label{sec:applications}

The interactive demonstration tool released alongside this paper enables several applications.

\textbf{Literary analysis.} The multi-resolution cluster view provides a new lens for close reading. The ``narrative timeline'' view for a single novel shows which concepts an author employs and where transitions between them occur---revealing structural patterns invisible to traditional reading.

\textbf{Comparative literature.} The cluster explorer reveals structural kinship between texts from different traditions. The cross-genre clusters documented in Section~\ref{sec:crossgenre}---where the same narrative beat appears in Gothic horror, domestic romance, religious fiction, and satirical comedy---provide empirical evidence for structural universals in narrative that literary theory has long hypothesised but struggled to demonstrate at scale.

\textbf{Education.} The concept labels and sample passages provide a data-grounded vocabulary for discussing narrative structure. Students can see that ``lyrical landscape meditation'' is a recurrent pattern employed by hundreds of authors across centuries, then examine how different authors deploy it.

\textbf{Corpus-scale analysis.} The $k = 2{,}000$ concept map provides a structured view of English-language literary production. The specificity of fine-grained clusters---Franco-Prussian War accounts, Darwin-Huxley correspondence, American founding father rhetoric---enables systematic study of literary traditions at a scale that would be impractical through manual reading.

\textbf{Beyond text.} The mechanism that produces transferable concepts here is not specific to literary text. Temporal co-occurrence under compression extracts recurrent patterns from any sequential data with sufficient regularity. The conditions are: enough independent sequences (thousands of books, in this case), recurring structural patterns across those sequences, and a capacity bottleneck that forces compression beyond memorisation. Any domain meeting these conditions is a candidate---legal case law organised by citation co-occurrence, gene expression profiles organised by regulatory co-occurrence, or user interaction sequences organised by behavioural co-occurrence. If confirmed in other domains, this would establish a general mechanism for unsupervised concept formation from sequential experience. We consider this the most significant implication of the present work: not the specific concepts discovered in literary text, but the demonstration that compression of temporal co-occurrence can produce transferable structural abstractions without supervision.

\section{Limitations}
\label{sec:limitations}

\textbf{Single training run.} All results come from single training runs (2K pilot and 10K full). Multi-seed evaluation has not been performed. The training accuracy (42.75\%) is a single data point, not a validated optimum.

\textbf{Temporal shuffle on pilot only.} The temporal shuffle control was performed on the 2K pilot corpus, not the full 10K corpus. While the same architecture was used, a shuffle control on the actual experimental data would be more rigorous.

\textbf{BGE baseline on subset.} The raw BGE clustering baseline was computed on a 2,000-novel subset, not the full corpus. A matched comparison on all 9,766 novels would be more rigorous.

\textbf{Cluster labels are post-hoc.} Concept labels are generated by an LLM examining sample passages, not by the model itself. They are interpretive aids, not outputs of the method.

\textbf{No systematic accuracy sweep.} The relationship between compression ratio and concept quality is hypothesised but not empirically validated across multiple accuracy levels.

\textbf{English language only.} The corpus is overwhelmingly English. Whether temporal co-occurrence discovers analogous transition-structure concepts in other languages is likely but untested.

\textbf{Gutenberg selection bias.} Project Gutenberg over-represents 19th and early 20th century English-language literature. The discovered concepts reflect this corpus, not literature in general.

\textbf{Chunk size not varied.} All results use 50-token chunks with 15-token overlap. The interaction between chunk size and $k$ value is unexplored.

\textbf{No formal human evaluation.} This work emphasises large-scale unsupervised structure discovery and released interactive inspection tooling. The claim that clusters capture transition structure relies on qualitative inspection and LLM-generated labels; formal human annotation of cluster coherence and category type is left to future work.

\textbf{No downstream task evaluation.} Unlike AAR, which demonstrates downstream QA improvement, we do not evaluate the discovered concepts on any downstream task. The concepts are presented as intrinsically interesting structure rather than as features useful for a specific application.

\section{Conclusion}
\label{sec:conclusion}

Temporal co-occurrence within texts discovers recurrent transition-structure concepts---patterns organised by what passages \emph{do} in context rather than what they \emph{say}. Training a contrastive model on which passages appear near which other passages, under compression that forces generalisation beyond individual co-occurrence pairs, produces a multi-resolution concept map spanning cross-author transition-structure patterns over 9,766 texts. These concepts span narrative functions, discourse registers, literary traditions, scene templates, and subject-matter conventions. They are qualitatively different from similarity-based groupings: more diverse in book coverage, more selective in unseen-novel assignment, and organised by structural role rather than semantic content.

The method requires no labels, no genre metadata, and no topic annotations. The training signal is temporal co-occurrence; the concepts emerge from compression. The same architecture and training signal that PAM uses for episodic recall and AAR uses for multi-hop retrieval here yields qualitatively different behaviour when the compression regime interacts with the regularity of the underlying patterns.

An interactive demonstration tool is released alongside this paper at \url{https://eridos.ai/concept-discovery}, enabling exploration of the full concept map across all six granularities for any text in the corpus, plus the five featured evaluation novels and five additional unseen novels. Code and data are available at \url{https://github.com/EridosAI/PAM-Concept-Discovery}.

\bibliographystyle{plainnat}
\bibliography{concept_references}

@article{dury2026pam,
  author    = {Dury, Jason},
  title     = {Predictive Associative Memory: Retrieval Beyond Similarity Through Temporal Co-occurrence},
  year      = {2026},
  journal   = {arXiv preprint arXiv:2602.11322},
  url       = {https://arxiv.org/abs/2602.11322},
}

@misc{dury2026aar,
  author    = {Dury, Jason},
  title     = {Association $\neq$ Similarity: Learning Corpus-Specific Associations for Multi-Hop Retrieval},
  year      = {2026},
  doi       = {10.5281/zenodo.18602385},
  url       = {https://doi.org/10.5281/zenodo.18602385},
  note      = {Zenodo preprint},
}

@article{blei2003lda,
  author  = {Blei, David M. and Ng, Andrew Y. and Jordan, Michael I.},
  title   = {Latent {Dirichlet} Allocation},
  journal = {Journal of Machine Learning Research},
  volume  = {3},
  pages   = {993--1022},
  year    = {2003},
}

@article{grootendorst2022bertopic,
  author  = {Grootendorst, Maarten},
  title   = {{BERTopic}: Neural Topic Modeling with a Class-Based {TF-IDF} Procedure},
  journal = {arXiv preprint arXiv:2203.05794},
  year    = {2022},
}

@inproceedings{chambers2008narrative,
  author    = {Chambers, Nathanael and Jurafsky, Dan},
  title     = {Unsupervised Learning of Narrative Event Chains},
  booktitle = {Proceedings of the 46th Annual Meeting of the Association for Computational Linguistics (ACL)},
  year      = {2008},
}

@article{reagan2016arcs,
  author  = {Reagan, Andrew J. and Mitchell, Lewis and Kiley, Dilan and Danforth, Christopher M. and Dodds, Peter Sheridan},
  title   = {The Emotional Arcs of Stories Are Dominated by Six Basic Shapes},
  journal = {EPJ Data Science},
  volume  = {5},
  number  = {1},
  pages   = {31},
  year    = {2016},
}

@misc{jockers2015syuzhet,
  author       = {Jockers, Matthew L.},
  title        = {Syuzhet: Extracts Sentiment and Sentiment-Derived Plot Arcs from Text},
  year         = {2015},
  note         = {CRAN R package},
}

@inproceedings{elson2010social,
  author    = {Elson, David K. and Dames, Nicholas and McKeown, Kathleen R.},
  title     = {Extracting Social Networks from Literary Fiction},
  booktitle = {Proceedings of the 48th Annual Meeting of the Association for Computational Linguistics (ACL)},
  year      = {2010},
}

@inproceedings{chambers2009schemas,
  author    = {Chambers, Nathanael and Jurafsky, Dan},
  title     = {Unsupervised Learning of Narrative Schemas and Their Participants},
  booktitle = {Proceedings of the Joint Conference of the 47th Annual Meeting of the ACL and the 4th International Joint Conference on Natural Language Processing (ACL-IJCNLP)},
  year      = {2009},
}

@book{propp1968morphology,
  author    = {Propp, Vladimir},
  title     = {Morphology of the Folktale},
  publisher = {University of Texas Press},
  year      = {1968},
  note      = {Originally published 1928},
}

@article{oord2018cpc,
  author  = {van den Oord, A{\"a}ron and Li, Yazhe and Vinyals, Oriol},
  title   = {Representation Learning with Contrastive Predictive Coding},
  journal = {arXiv preprint arXiv:1807.03748},
  year    = {2018},
}

@inproceedings{radford2021clip,
  author    = {Radford, Alec and Kim, Jong Wook and Hallacy, Chris and Ramesh, Aditya and Goh, Gabriel and Agarwal, Sandhini and Sastry, Girish and Askell, Amanda and Mishkin, Pamela and Clark, Jack and Krueger, Gretchen and Sutskever, Ilya},
  title     = {Learning Transferable Visual Models from Natural Language Supervision},
  booktitle = {Proceedings of the 38th International Conference on Machine Learning (ICML)},
  year      = {2021},
}

@inproceedings{gao2021simcse,
  author    = {Gao, Tianyu and Yao, Xingcheng and Chen, Danqi},
  title     = {{SimCSE}: Simple Contrastive Learning of Sentence Embeddings},
  booktitle = {Proceedings of the 2021 Conference on Empirical Methods in Natural Language Processing (EMNLP)},
  year      = {2021},
}

@article{izacard2022contriever,
  author  = {Izacard, Gautier and Caron, Mathilde and Hosseini, Lucas and Rber, Sebastian and Grave, Armand and Bojanowski, Piotr and Joulin, Armand},
  title   = {Unsupervised Dense Information Retrieval with Contrastive Learning},
  journal = {Transactions on Machine Learning Research},
  year    = {2022},
}

@inproceedings{karpukhin2020dpr,
  author    = {Karpukhin, Vladimir and O\u{g}uz, Barlas and Min, Sewon and Lewis, Patrick and Wu, Ledell and Edunov, Sergey and Chen, Danqi and Yih, Wen-tau},
  title     = {Dense Passage Retrieval for Open-Domain Question Answering},
  booktitle = {Proceedings of the 2020 Conference on Empirical Methods in Natural Language Processing (EMNLP)},
  year      = {2020},
}

@inproceedings{xiao2024bge,
  author    = {Xiao, Shitao and Liu, Zheng and Zhang, Peitian and Muennighoff, Niklas},
  title     = {{C-Pack}: Packaged Resources to Advance General {Chinese} Embedding},
  booktitle = {Proceedings of the 47th International ACM SIGIR Conference on Research and Development in Information Retrieval (SIGIR)},
  year      = {2024},
}

@misc{baai2023bge,
  author       = {{BAAI}},
  title        = {{BAAI/bge-large-en-v1.5}},
  year         = {2023},
  howpublished = {Hugging Face model card, \url{https://huggingface.co/BAAI/bge-large-en-v1.5}},
}

@article{mcclelland1995cls,
  author  = {McClelland, James L. and McNaughton, Bruce L. and O'Reilly, Randall C.},
  title   = {Why There Are Complementary Learning Systems in the Hippocampus and Neocortex: Insights from the Successes and Failures of Connectionist Models of Learning and Memory},
  journal = {Psychological Review},
  volume  = {102},
  number  = {3},
  pages   = {419--457},
  year    = {1995},
}

@article{wilson1994reactivation,
  author  = {Wilson, Matthew A. and McNaughton, Bruce L.},
  title   = {Reactivation of Hippocampal Ensemble Memories During Sleep},
  journal = {Science},
  volume  = {265},
  number  = {5172},
  pages   = {676--679},
  year    = {1994},
}

@misc{gutenberg,
  author       = {{Project Gutenberg}},
  title        = {Project {Gutenberg}},
  howpublished = {\url{https://www.gutenberg.org/}},
  year         = {2024},
  note         = {Accessed 2026},
}

@misc{gutendex,
  author       = {Johnson, Garth},
  title        = {Gutendex: A {JSON} Web {API} for {Project Gutenberg} Ebook Metadata},
  howpublished = {\url{https://gutendex.com/}},
  year         = {2024},
}

\end{document}